\documentclass[runningheads]{llncs}
\usepackage{graphicx}
\usepackage{float}
\usepackage{mathtools}
\usepackage[bottom]{footmisc}

\DeclarePairedDelimiter{\norm}{\lVert}{\rVert}

\begin{document}
\title{Improving the Segmentation of Anatomical Structures in Chest Radiographs using U-Net with an ImageNet Pre-trained Encoder}
%
%
\author{Maayan Frid-Adar\inst{1} \and
Avi Ben-Cohen\inst{2} \and
Rula Amer \inst{2} \and
Hayit Greenspan\inst{1,2}}
\authorrunning{Frid-Adar et al.}
\titlerunning{Improving the Segmentation of Anatomical Structures}
%
\institute{RADLogics Ltd., Tel-Aviv, Israel \and
Tel Aviv University, Faculty of Engineering, Department of Biomedical Engineering, Medical Image Processing Laboratory, Tel Aviv 69978, Israel} 
%
\maketitle{}             
\begin{abstract}
Accurate segmentation of anatomical structures in chest radiographs is essential for many computer-aided diagnosis tasks. In this paper we investigate the latest fully-convolutional architectures for the task of multi-class segmentation of the lungs field, heart and clavicles in a chest radiograph.
In addition, we explore the influence of using different loss functions in the training process of a neural network for semantic segmentation.
We evaluate all models on a common benchmark of 247 X-ray images from the JSRT database and ground-truth segmentation masks from the SCR dataset.
Our best performing architecture, is a modified U-Net that benefits from pre-trained encoder weights. This model outperformed the current state-of-the-art methods tested on the same benchmark, with Jaccard overlap scores of 96.1\% for lung fields, 90.6\% for heart and 85.5\% for clavicles.

\keywords{Chest radiographs  \and Lung segmentation \and Clavicle segmentation \and Heart segmentation \and Fully convolutional networks}

\end{abstract}
\section{Introduction}
Approximately 3.6 billion diagnostic radiological examinations, such as radiographs (x-rays), are performed globally every year \cite{report}. Chest radiographs are performed to evaluate the lungs, heart and thoracic viscera. They are crucial for diagnosing various lung disorders in all levels of health care.
Computer-aided diagnostic (CAD) tools serve an important role to assist the radiologists with the growing number of chest radiographs. Accurate segmentation of anatomical structures in chest radiographs is essential for many analysis tasks in CAD. For example: segmentation of the lungs field can help detecting lung diseases and shape irregulars; segmentation of the heart outline  can help to predict cardiomegaly; and the segmentation of clavicles  can improve the diagnosis of pathologies near the apex of the lung.

Evaluating a chest radiograph is a challenging task due to the high variability between patients, unclear and overlapping organs borders, and image artifacts. A clear and high quality radiograph is not easy to acquire.
This challenge drew many researchers over the years to improve the segmentation of anatomical structures in chest radiographs \cite{novikov,park,ibragimov,yang}. An open benchmark dataset that was provided by Ginneken et al. \cite{ginneken_scr} facilitated over the years an objective comparison between the different segmentation methods. Classic approaches include active shape and appearance models, pixel classification methods, hybrid models and landmark based models. More recently deep learning approaches were suggested \cite{novikov,park} based on the successful employment of convolutional neural networks (CNNs) on various detection and segmentation tasks in the medical imaging domain \cite{dl_overview}.

CNN architectures for semantic segmentation usually incorporate encoder and decoder networks \cite{unet_olaf,FCN} that reduce the resolution of the image to capture the most important details and then restore the resolution of the image. Another semantic segmentation approach is to keep the resolution of the network by incorporating dilated convolutions \cite{DRN} that enlarge the global receptive field of the CNN to larger context information. In both approaches, the CNN can output  single-class or multiple-class segmentation masks. The resolution of the output mask is the same as the input radiograph image. The training process of each CNN is affected by several training features: One is the selection of the loss function that guides the optimization process during the training process (with different loss functions effecting differently  the final output segmentation performance results); The other is the initialization of the network weights - random initialization or weights transferred from another trained network (transfer learning from a totally different task).

In this paper, we explore the segmentation of anatomical structures in chest radiographs, namely the lungs field, the heart and the clavicles, using a set of the most advanced CNN architectures for multi-class semantic segmentation. 
We propose an improved encoder-decoder style CNN with pre-trained weights of the encoder network and show its superiority over other state of the art CNN architectures. We further examine the use of multiple loss functions for training the best selected network and the effect of multi-class vs. single-class training.
We present qualitative and quantitative comparisons on a common benchmark data, based on the JSRT database \cite{JSRT}. Our best performing model, the U-net with an ImageNet pre-trained encoder, outperformed the currently state-of-the-art segmentation methods for all anatomical structures.

\section{Methods}
\subsection{Fully Convolutional Neural Network Architectures}
\label{FCN_arch}
Fully convolutional networks (FCN) are extensively used for semantic segmentation tasks. In this study, four different state of the art architectures have been tested as follows:

\textbf{FCN} - The first FCN architecture that we used in this work is based on the FCN-8s net that uses the VGG-16 layer net \cite{FCN,vgg}. The VGG-16 net is converted into an FCN by decapitating the final classification layer and converting fully connected layers into convolution. Deconvolution layers are then used to upsample the coarse outputs to pixel-dense outputs. Skip connections are used to merge output from previous pooling layers in the network which was shown to improve the segmentation quality \cite{FCN}. 

\textbf{Fully convolutional DenseNet} - The second network architecture that was tested is based on the fully convolutional DenseNet shown in \cite{fc_densenet_tiramisu}. DenseNet architecture \cite{densenet} proposes intensive layer fusion. Each dense block consists of a set of convolution layers using a similar scale where each convolution layer processes the concatenation of all its previous layers thus enabling the fusion of  numerous representation levels. For the fully convolutional DenseNet architecture a decoding path is added to generate the segmentation output. The fusion between different layers consists of intra dense block layers fusion as well as the concatenation of the preceding high level feature maps and the ones coming from the encoding block at the same scale.

\textbf{Dilated residual networks} - The dilated residual network (DRN) \cite{DRN} uses dilated convolution \cite{atrous} to increase the resolution of output feature maps without reducing the receptive field of individual neurons. It was shown to improve the performance compared to the standard residual networks presented in \cite{resnet}. We have implemented the DRN-C-26 as stated in \cite{DRN}.

\begin{figure}
\centering
\includegraphics[height=5.0 cm]{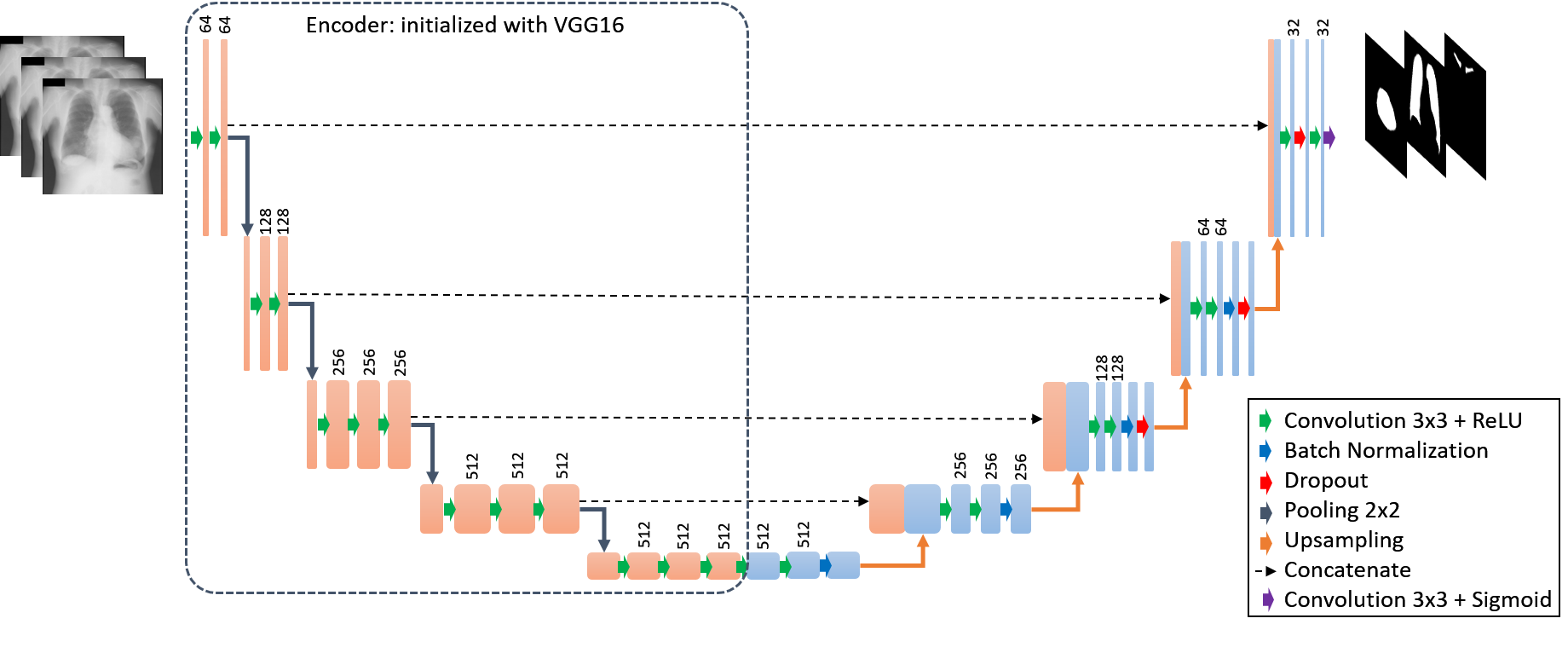}
\caption{The proposed U-Net architecture with a VGG-16 based encoder.}
\label{fig:architecture}\end{figure}

\textbf{U-Net with VGG-16 encoder} - The U-Net architecture \cite{unet_olaf} has been extensively used for different image-to-image tasks in computer vision with a major contribution to the image segmentation task. The U-Net includes a contracting path (the encoder) with several layers of convolution and pooling for down-sampling. The second half of the network includes an expansion path (the decoder) that uses up-sampling and convolution layers sequentially to generate an output with a similar size as the input image. Additionally, the U-Net architecture combines the encoder features with the decoder features in different levels of the network using skip connections. Iglovikov et al \cite{TernausNet} proposed to use a VGG11 \cite{vgg} as an encoder which was pre-trained on ImageNet \cite{Imagenet} dataset and showed that it can improve the standard U-Net performance in binary segmentation of buildings in aerial images. A similar concept was used in the current study with the more advanced VGG16 \cite{vgg} as an encoder. Figure \ref{fig:architecture} shows a diagram of our proposed network. The chest X-ray image is duplicated to obtain an input image with 3 channels similar to the RGB images that are used as input to the VGG-16 net (which is the encoder in the proposed architecture). 

\subsection{Objective loss functions}
The loss function is used to guide the training process of a convolutional network by measuring the compatibility between the network prediction and the ground truth label. Let us denote S as the estimated segmentation mask and G
as the ground truth mask. In a multi-class semantic segmentation task including $C = \{c_1,...,c_m\}$ classes, the total loss (TS) between S and G is defined as the sum of losses in every class:
\begin{equation}
TL(S,G)=\sum_{c=1}^{m}L_c(S,G)
\end{equation}
In this study we explore the influence of using different loss functions in the FCNs training process. The Dice similarity coefficient (DSC) and Jaccard similarity coefficient (JSC) are two well known measures in segmentation and can be used as objective loss functions in training. These segmentation measures between S and G are defined as:
\begin{equation}
DSC(S,G)=2\frac{|SG|}{|S|+|G|}
\end{equation}
\begin{equation}
JSC(S,G)=\frac{|SG|}{|S|+|G|-|SG|}
\end{equation}
when used as loss in training, both measures weights FP and FN detections equally. The Tversky loss \cite{Tversky} introduces weighting into the loss function for highly imbalanced data, where we want to segment small objects. 
The Tversky index is defined as:
\begin{equation}
Tversky(S,G;\alpha,\beta)=\frac{|SG|}{|SG|+\alpha|S / G|+\beta|G / S|}
\end{equation}
where $\alpha$ and $\beta$ control the magnitude of penalties for FPs and FNs, respectively. In our study we used $\alpha=0.3$ and $\beta=0.7$.

An additional loss function tested is the Binary Cross-Entropy (BCE). BCE was calculated separately for each class segmentation map. For each pixel $s_i\in S$ and pixel $g_i\in G$ that share the same pixel position i, the loss is averaged over all pixels $N$ as follows:
\begin{equation}
BCE(S,G)=\frac{1}{N}\sum_{i=1}^N g_i\log(s_i) + (1-g_i)\log(1-s_i)
\end{equation}

\section{Segmentation of Anatomical Structures}
\subsection{Dataset}
\begin{figure} [H]
\centering
\includegraphics[height=3.0 cm]{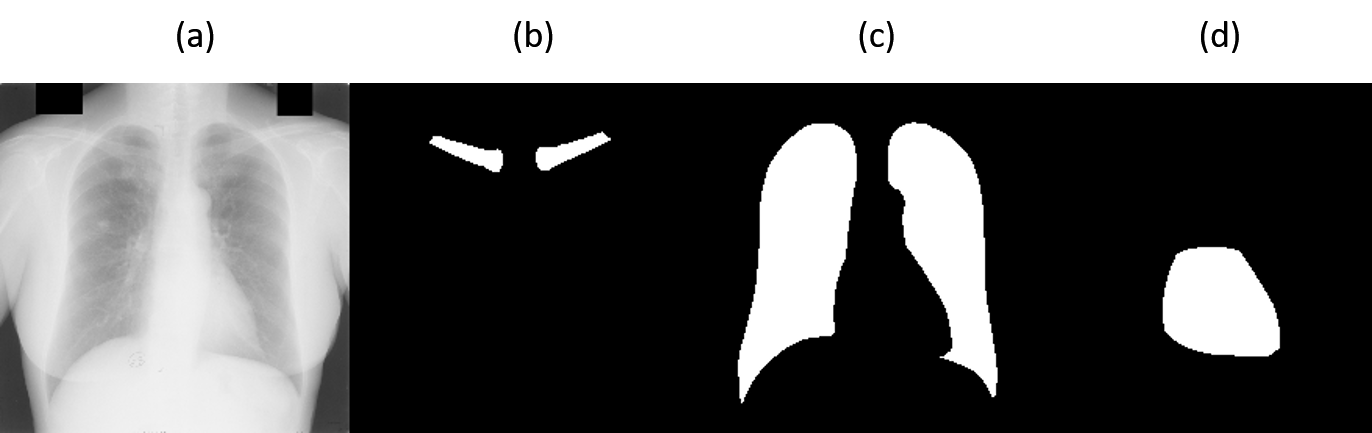}
\caption{Data sample from \cite{ginneken_scr}: (a) chest radiograph image; (b) clavicles segmentation mask; (c) lung segmentation mask; (d) heart segmentation mask.}
\label{fig:data_sample}
\end{figure}

Evaluation of the chest anatomical structures segmentation was done on chest radiographs from the JSRT database \cite{JSRT}. This public database includes 247 posterior-anterior (PA) chest radiograph images of size $2048\times2048$ pixels, 0.175 mm pixel spacing and 12-bit gray levels. Ginneken et al. \cite{ginneken_scr} publicized the Segmentation in Chest Radiographs (SCR) database, a benchmark set of segmentation masks for the lungs field, heart and clavicles (see Figure \ref{fig:data_sample}). The annotations were made by two human observers and a radiologist consultant. The segmentations of the first observer generate the ground-truth segmentation masks and the other -  human observer results. The benchmark data is split into two folds of 124 and 123 cases, each containing equal amount of normal cases and cases with lung nodules. Following the suggested instructions for comparison between the segmentation results, images in one fold were used for training and images from the other fold were used for testing, and vise versa.
The final evaluation is defined as the average performance over the two folds.

For training, we resize the images to $224\times 224$ pixels and normalize each image by its mean and standard deviation. The networks are trained using Adam optimizer with initial learning rate of $10^{-5}$ and default parameters for 100 epochs.
We use augmentations of scaling, translation and small rotations.
In testing, We threshold the output score maps with $threshold = 0.25$ to generate binary segmentation masks of each anatomical structure.

\subsection{Performance Measures}
To measure the performance of the proposed architectures and compare to state-of-the-art results, we use well accepted metrics for segmentation: Dice similarity coefficient, Jaccard index (also known as intersection over union) and mean absolute contour distance (MACD).
MACD is a measure of distance between two contours. For each point on contour A, the closest point on contour B is computed by the euclidean distance $d(a_{i},B) = min_{b_{j}\in B}\norm{b_{j} - a_{i}}$. The distance values are then averaged over all points. Since distances from A to B are not the same as B to A, we derive a common average between the two averages as follows:
\begin{equation}
MACD(A,B)=\frac{1}{2}(\frac{\sum_{i=1}^{n} d(a_{i},B)}{n} + \frac{\sum_{i=1}^{m} d(b_{i},A)}{m})
\end{equation}
Because MACD measure is given in millimeters, we multiply the original pixel spacing by a factor of $2048 / {224}$ to match the target image resolution.

\subsection{Experimental Results}
Table \ref{tabel:arch_compare} compares the segmentation performance of the four state of the art fully convolutional networks for semantic segmentation as listed in section \ref{FCN_arch}. All models are trained for multi-class segmentation into three classes: $lungs \ field,\\
heart, \ clavicles$. We use the $sigmoid$ activation function after the last layer of each network with $Dice$ as the loss function.
An additional column in Table \ref{tabel:arch_compare} shows if the network is fine-tunned (FT) from a pre-trained network.

The results show that the best performing architecture for the segmentation of all anatomical structures in chest radiograph, is the U-Net including the VGG16 encoder pre-trained on ImageNet. This architecture achieved the highest segmentation overlap scores (Jaccard) of 0.961, 0.906 and 0.855 for the Lungs field, Heart and Clavicles respectively.
It is noticeable that between all four architectures, the fine-tuned networks performed better than the networks trained from scratch.

\begin{table}[t]
\caption{Segmentation results of four compared architectures trained with multi-class Dice loss showing the Dice(D), Jaccard (J) and MACD metrics. Fine tuned (FT) architectures include a pre-trained VGG16 as an initial encoder.}
\label{tabel:arch_compare}
\centering
\begin{tabular}{|c|c|c|c|c|c|c|c|c|c|c|}
\cline{3-11}
\multicolumn{2}{c}{} &\multicolumn{3}{|c|}{Lungs} & \multicolumn{3}{c|}{Heart} & \multicolumn{3}{c|}{Clavicles} \\
\hline
Architecture & FT & D & J & MACD & D & J & MACD & D & J & MACD \\
\hline
\hline
FCN  & v & 0.976 & 0.953 & 1.341 & 0.944 & 0.895 & 3.099 & 0.884 & 0.795 & 1.277 \\
\hline
U-Net (VGG16) & v & \textbf{0.980} & \textbf{0.961} & \textbf{1.121} & \textbf{0.950} & \textbf{0.906} & \textbf{2.569} & \textbf{0.921} & \textbf{0.855} & \textbf{0.871} \\
\hline
FC DenseNet  & {} & 0.973 & 0.947 & 1.511 & 0.934 & 0.879 & 3.396 & 0.884 & 0.796 & 1.349 \\
\hline
DRN  & {} & 0.966 & 0.935 & 1.842 & 0.936 & 0.881 & 3.365 & 0.840 & 0.727 & 1.860 \\
\hline
\end{tabular}
\end{table}

\begin{figure}[h]
\centering
\includegraphics[height=6.5 cm]{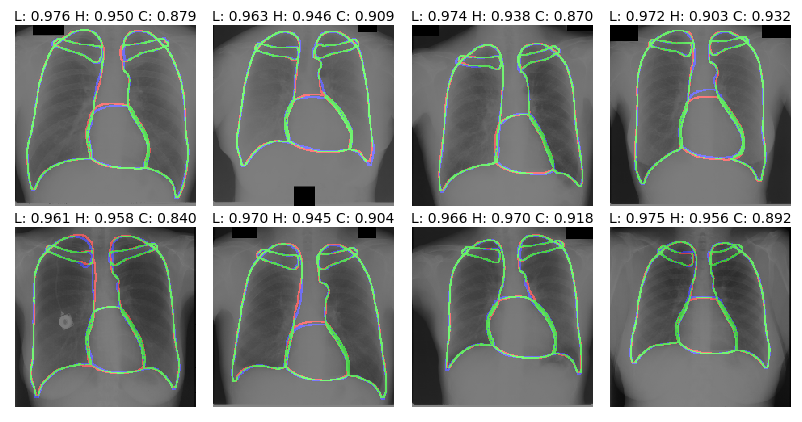}
\caption{Segmentation results of our best performing architecture with Jaccard score above each image for the Lungs(L), Heart(H) and Clavicles(C); Ground-truth segmentation is shown in blue, CNN segmentation in red and the overlap (true detections) in green.}
\label{fig:examples}
\end{figure}

For the top performing architecture, the U-Net based network, we further analyzed several training features.
Table \ref{tabel:loss_compare} summarizes the multi-class segmentation performance using different objective loss functions. It is evident that structures with smaller pixel area, like the clavicles, benefits from loss metrics with pixel weighing such as Tversky loss function.  
We also tested the performance of training a single-class network for each of the three classes vs. the multi-class training. For the lungs, the single class training did not resolve in significant improvement. However, for the heart and clavicles, the Dice and Jaccard scores in a single-class training were improved each by 1\% in comparison to the multi-class training.
The last improvement in performance of the multi-class segmentation was achieved using post-processing including small objects removal and hole fill. While the Dice and Jaccard metrics were not improved, the MACD metric showed an improvement from 1.121, 2.569 and 0.871 [mm] for the lungs, heart and clavicles to 1.019, 2.549 and 0.856 [mm] respectively. Figure \ref{fig:examples} shows a few segmentation examples of our best performing model.
A comparison of our U-Net based model trained with multi-class dice loss to existing state-of-the-art methods, validated on the same benchmark of chest radiographs and a human observer, is presented in Table \ref{tabel:state_of_the_art_compare}.

\begin{table}[t]
\caption{Multi-class segmentation results using different loss functions including DSC, JSC, Tversky and BCE (rows). The Dice(D), Jaccard (J) and MACD are used as metrics (columns) for each anatomical structure.}
\label{tabel:loss_compare}
\centering
\begin{tabular}{|c|c|c|c|c|c|c|c|c|c|}
\cline{2-10}
\multicolumn{1}{c}{} &\multicolumn{3}{|c|}{Lungs} & \multicolumn{3}{c|}{Heart} & \multicolumn{3}{c|}{Clavicles} \\
\hline
Loss Function & D & J & MACD & D & J & MACD & D & J & MACD \\
\hline
\hline
DSC & \textbf{0.980} & \textbf{0.961} & 1.121 & \textbf{0.950} & \textbf{0.906} & \textbf{2.569} & 0.921 & 0.855 & \textbf{0.871} \\
\hline
JSC & 0.979 & 0.960 & \textbf{1.082} & 0.949 & 0.905 & 2.602 & 0.921 & 0.855 & 0.920 \\
\hline
Tversky & 0.979 & 0.960 & 1.139 & \textbf{0.950} & 0.905 & 2.581 & \textbf{0.923} & \textbf{0.858} & 0.987 \\
\hline
BCE &  \textbf{0.980} & \textbf{0.961} & 1.119 & \textbf{0.950} & \textbf{0.906} & 2.592 & 0.911 & 0.838 & 1.145 \\
\hline
\end{tabular}
\end{table}

\begin{table}[h]
\caption{Our best performing architecture compared to state-of-the-art models; "-" means that the score was not reported; (*) used different data split than suggested in SCR benchmark}
\label{tabel:state_of_the_art_compare}
\centering
\begin{tabular}{p{4cm} p{2.5cm} p{2.5cm} p{2.5cm}}
\hline
{} & Dice & Jaccard & MACD (mm) \\
\hline
\textit{Lungs} & {} & {} & {} \\
Human observer \cite{ginneken_scr} & - & $0.946\pm 0.018$ & $1.64\pm 0.69$ \\
Hybrid voting \cite{ginneken_scr} & - & $0.949\pm 0.020$ & $1.62\pm 0.66$ \\
Ibragimov et al. \cite{ibragimov} & - & $0.953\pm 0.020$ & $1.43\pm 0.85$ \\
Hwang and Park \cite{park} & $0.980\pm 0.008$ & $0.961\pm 0.015$ & $1.237\pm 0.702$ \\
Novikov et al. \cite{novikov}(*) & $0.974$ & $0.950$ & - \\
Yang et al. \cite{yang} & $0.975\pm 0.001$ & $0.952\pm 0.018$ & $1.37\pm 0.67$ \\
U-Net (VGG16) & $0.980\pm 0.008$ & $0.961\pm 0.014$ & $1.019\pm 0.564$ \\
\\[-0.7em]
\textit{Heart} & {} & {} & {} \\
Human observer \cite{ginneken_scr} & - & $0.878\pm 0.054$ & $3.78\pm 1.82$ \\
Hybrid voting \cite{ginneken_scr} & - & $0.860\pm 0.056$ & $4.24\pm 1.87$ \\
Novikov et al. \cite{novikov}(*) & $0.937$ & $0.882$ & - \\
U-Net (VGG16) & $0.950\pm 0.021$ & $0.906\pm 0.038$ & $2.549\pm 1.126$ \\
\\[-0.7em]
\textit{Clavicles} & {} & {} & {} \\
Human observer \cite{ginneken_scr} & - & $0.896\pm 0.037$ & $0.68\pm 0.26$ \\
Hybrid voting \cite{ginneken_scr} & - & $0.736\pm 0.106$ & $1.88\pm 0.93$ \\
Novikov et al. \cite{novikov}(*) & $0.929$ & $0.868$ & - \\
U-Net (VGG16) & $0.921\pm 0.027$ & $0.855\pm 0.045$ & $0.855\pm 0.322$ \\
\hline
\end{tabular}
\end{table}

\section{Discussion and Conclusion}
Segmentation of anatomical structures in chest radiographs is a challenging task that attracted  considerable interest over the years. The advantages of newly introduced CNN architectures, together with the public benchmark dataset provided in \cite{ginneken_scr} on the JSRT images, motivated further studies  in this field. Some of the recent studies focused only on the problem of lung segmentation, and a few have also dealt with the problem of heart and clavicles segmentation.
In this paper, we employed and evaluated the segmentation performance of four top FCN architectures \cite{fc_densenet_tiramisu,FCN,DRN,TernausNet} for semantic segmentation for all three anatomical structures, using multi-class dice loss. 

The network architectures presented in this study are well known and showed promising results in many computer vision semantic segmentation tasks. The FCN \cite{FCN} and the U-Net \cite{unet_olaf} are considered classical approaches while the FC DenseNet and the DRN are more advanced and relatively new approaches for semantic segmentation. Hence, it was interesting to see in Table \ref{tabel:arch_compare} that the classic U-Net and FCN showed superior segmentation performance over the more advanced approaches. The advantage of using pre-trained networks for medical imaging tasks has already been shown in several studies \cite{dl_overview}, and even though only the encoder part of the FCN and U-Net (VGG16 encoder) networks was pre-trained using the ImageNet database in our case, it seemed to be advantageous. The best segmentation performance was obtained using the proposed U-Net based architecture including the pre-trained VGG16 encoder (Table \ref{tabel:arch_compare}).

Next, we explored the effect of training multi-class segmentation model using different loss functions (Table \ref{tabel:loss_compare}). We demonstrated that small structures such as the clavicles can benefit from  weighted loss functions such the Tversky loss function while the larger structures (lung and heart) achieved the best segmentation results using Dice or Binary Cross-Entropy loss functions.  Applying  additional minor post-processing resulted in further decrease of the MACD measure with cleaner and more precise segmentations for all three structures as displayed in Figure \ref{fig:examples}. 

Table \ref{tabel:state_of_the_art_compare} presents the final comparison between our top selected model, the multi-class U-Net VGG16 with dice loss, to state-of-the-art methods \cite{ginneken_scr,ibragimov,park,novikov,yang} and  human observer segmentations \cite{ginneken_scr}. Our model outperformed all state-of-the-art methods tested in this study and the human observer for the lungs and heart segmentation. For the clavicles segmentation, fewer studies were conducted. Novikov et al. \cite{novikov} reported results on different data split than the benchmark recommendation so its not an objective comparison. However, our proposed network outperformed an additional top reported method \cite{ginneken_scr}.

In conclusion, we presented an experimental study in which four top segmentation architectures and several losses were compared for the task of segmenting anatomical structures on chest X-Ray images. Results were evaluated quantitatively with qualitative examples of our best performing model. Improving the segmentation of the lung field, heart and clavicles is the foundation for better CAD tools and the development of new applications for medical thoracic images analysis.

%
%
%

\begin{thebibliography}{8}


\bibitem{report}
United Nations. Scientific Committee on the Effects of Atomic Radiation. Report of the United Nations Scientific Committee on the Effects of Atomic Radiation: Fifty-sixth Session (10-18 July 2008) (No. 46). United Nations Publications. (2008).

\bibitem{novikov}
Novikov, A. A., Lenis, D., Major, D., Hladůvka, J., Wimmer, M., and Bühler, K.: Fully convolutional architectures for multi-class segmentation in chest radiographs. IEEE Transactions on Medical Imaging, (2018).

\bibitem{park}
Hwang, S., and Park, S.: Accurate Lung Segmentation via Network-Wise Training of Convolutional Networks. In Deep Learning in Medical Image Analysis and Multimodal Learning for Clinical Decision Support (pp. 92-99). Springer, Cham. (2017).

\bibitem{ibragimov}
Ibragimov, B., Likar, B., Pernuš, F., and Vrtovec, T.: Accurate landmark-based segmentation by incorporating landmark misdetections. In Biomedical Imaging (ISBI), 2016 IEEE 13th International Symposium on (pp. 1072-1075). IEEE, (2016).

\bibitem{yang}
Yang, W., Liu, Y., Lin, L., Yun, Z., Lu, Z., Feng, Q., and Chen, W.: Lung field segmentation in chest radiographs from boundary maps by a structured edge detector. IEEE journal of biomedical and health informatics, 22(3), 842-851. (2018). 

\bibitem{ginneken_scr}
Van Ginneken, B., Stegmann, M. B., and Loog, M.: Segmentation of anatomical structures in chest radiographs using supervised methods: a comparative study on a public database. Medical image analysis, 10(1), 19-40, (2006).

\bibitem{dl_overview}
Greenspan, H., van Ginneken, B., and Summers, R. M.: Guest editorial deep learning in medical imaging: Overview and future promise of an exciting new technique. IEEE Transactions on Medical Imaging, 35(5), 1153-1159. (2016).

\bibitem{unet_olaf}
Ronneberger, O., Fischer, P. and Brox, T.: U-net: Convolutional networks for biomedical image segmentation. In International Conference on Medical image computing and computer-assisted intervention (pp. 234-241). Springer, Cham. (2015).

\bibitem{FCN}
Long, J., Shelhamer, E., and Darrell, T.: Fully convolutional networks for semantic segmentation. In Proceedings of the IEEE conference on computer vision and pattern recognition (pp. 3431-3440). (2015).

\bibitem{DRN}
Yu, F., Koltun, V., and Funkhouser, T.: Dilated residual networks. In Computer Vision and Pattern Recognition (Vol. 1). (2017).

\bibitem{JSRT}
Shiraishi, J., Katsuragawa, S., Ikezoe, J., Matsumoto, T., Kobayashi, T., Komatsu, K. I., ... and Doi, K.: Development of a digital image database for chest radiographs with and without a lung nodule: receiver operating characteristic analysis of radiologists' detection of pulmonary nodules. American Journal of Roentgenology, 174(1), 71-74. (2000).

\bibitem{vgg}
Simonyan, K. and Zisserman, A.: Very deep convolutional networks for large-scale image recognition. arXiv preprint arXiv:1409.1556, (2014).

\bibitem{fc_densenet_tiramisu}
Jégou, S., Drozdzal, M., Vazquez, D., Romero, A., and Bengio, Y.: The one hundred layers tiramisu: Fully convolutional densenets for semantic segmentation. In Computer Vision and Pattern Recognition Workshops (CVPRW), IEEE Conference on (pp. 1175-1183). IEEE, (2017).

\bibitem{densenet}
Huang, G., Liu, Z., Weinberger, K. Q., and van der Maaten, L.: Densely connected convolutional networks. In Proceedings of the IEEE conference on computer vision and pattern recognition (Vol. 1, No. 2, p. 3), (2017).

\bibitem{atrous}
Chen, L. C., Papandreou, G., Kokkinos, I., Murphy, K., and Yuille, A. L.: Deeplab: Semantic image segmentation with deep convolutional nets, atrous convolution, and fully connected crfs. IEEE transactions on pattern analysis and machine intelligence, 40(4), 834-848, (2018).

\bibitem{resnet}
He, K., Zhang, X., Ren, S., and Sun, J.: Deep residual learning for image recognition. In Proceedings of the IEEE conference on computer vision and pattern recognition (pp. 770-778), (2016).

\bibitem{TernausNet}
Iglovikov, V. and Shvets, A.: TernausNet: U-Net with VGG11 Encoder Pre-Trained on ImageNet for Image Segmentation. arXiv preprint arXiv:1801.05746, (2018).

\bibitem{Imagenet}
Deng, J., Dong, W., Socher, R., Li, L.J., Li, K. and Fei-Fei, L.: Imagenet: A large-scale hierarchical image database. In Computer Vision and Pattern Recognition. CVPR 2009. IEEE Conference on (pp. 248-255). IEEE, (2009).

\bibitem{Tversky}
Salehi, S. S. M., Erdogmus, D., and Gholipour, A.: Tversky loss function for image segmentation using 3D fully convolutional deep networks. In International Workshop on Machine Learning in Medical Imaging (pp. 379-387). Springer, Cham. (2017).






\end{thebibliography}
%

\end{document}